\documentclass[10pt,conference,letterpaper]{IEEEtran}
\usepackage{array,multirow}
\usepackage{graphicx} 
\usepackage{grffile}
\usepackage{filecontents}
\usepackage{enumitem} 
\usepackage{amsmath}
\usepackage{amssymb}
\usepackage{nth}
\usepackage{booktabs}
\usepackage{amssymb}
\usepackage{xspace}
\usepackage{lastpage, flushend}
\usepackage{listings}
\usepackage{color}
\usepackage{pgfplots}
\usepackage{caption}
\usepackage{subcaption}
\usepackage{mwe}
\usepackage{multicol}
\usetikzlibrary{positioning,shadows}
\newcommand{\latinphrase}[1]{\textit{#1}}  
\newcommand{\etal}{\latinphrase{et~al.}\xspace}

\usepackage[hyphens]{url}
\makeatletter
\def\url@leostyle{%
	\@ifundefined{selectfont}{\def\UrlFont{\small\rmfamily}}%
	{\def\UrlFont{\rmfamily}}%
}
\makeatother	
\definecolor{darkgreen}{RGB}{47,109,79}
\definecolor{darkblue}{RGB}{57,79,99}
\definecolor{rosso}{RGB}{220,57,18}
\definecolor{giallo}{RGB}{255,153,0}
\definecolor{blu}{RGB}{102,140,217}
\definecolor{verde}{RGB}{16,150,24}
\definecolor{viola}{RGB}{153,0,153}
\definecolor{awesome}{rgb}{1.0, 0.13, 0.32}
\definecolor{ref}{rgb}{0.65,0.65,0.65} 

\tikzstyle{chart}=[
legend label/.style={font={\scriptsize},anchor=west,align=left},
legend box/.style={rectangle, draw, minimum size=5pt},
axis/.style={black,semithick,->},
axis label/.style={anchor=east,font={\tiny}},
]

\tikzstyle{bar chart}=[
chart,
bar width/.code={
	\pgfmathparse{##1/2}
	\global\let\bar@w\pgfmathresult
},
bar/.style={very thick, draw=white},
bar label/.style={font={\bf\small},anchor=north},
bar value/.style={font={\footnotesize}},
bar width=.75,
]

\tikzstyle{pie chart}=[
chart,
slice/.style={line cap=round, line join=round, very thick,draw=white},
pie title/.style={font={\bf}},
slice type/.style 2 args={
	##1/.style={fill=##2},
	values of ##1/.style={}
}
]

\pgfdeclarelayer{background}
\pgfdeclarelayer{foreground}
\pgfsetlayers{background,main,foreground}

\newcommand{\pie}[3][]{
	\begin{scope}[#1]
		\pgfmathsetmacro{\curA}{90}
		\pgfmathsetmacro{\r}{1}
		\def\c{(0,0)}
		\node[pie title] at (90:1.3) {#2};
		\foreach \v/\s in{#3}{
			\pgfmathsetmacro{\deltaA}{\v/100*360}
			\pgfmathsetmacro{\nextA}{\curA + \deltaA}
			\pgfmathsetmacro{\midA}{(\curA+\nextA)/2}
			
			\path[slice,\s] \c
			-- +(\curA:\r)
			arc (\curA:\nextA:\r)
			-- cycle;
			\pgfmathsetmacro{\d}{max((\deltaA * -(.5/50) + 1) , .5)}
			
			\begin{pgfonlayer}{foreground}
				\path \c -- node[pos=\d,pie values,values of \s]{$\v\%$} +(\midA:\r);
			\end{pgfonlayer}
			
			\global\let\curA\nextA
		}
	\end{scope}
}

\newcommand{\legend}[2][]{
	\begin{scope}[#1]
		\path
		\foreach \n/\s in {#2}
		{
			++(0,-10pt) node[\s,legend box] {} +(5pt,0) node[legend label] {\n}
		}
		;
	\end{scope}
}
\usetikzlibrary{patterns}

\title{Privacy-Preserving Deep Inference for Rich User Data on The Cloud}
\author{%

	Seyed Ali Osia{\small $~^{\sharp}$}, Ali Shahin Shamsabadi{\small $~^{\sharp}$}, Ali Taheri{\small $~^{\sharp}$}, Kleomenis Katevas{\small $~^{\star}$},  \\
	Hamid R. Rabiee{\small $~^{\sharp}$}, Nicholas D. Lane{\small $~^{\dagger}$}, Hamed Haddadi{\small $~^{\star}$} %
\vspace{1.6mm}\\
\fontsize{10}{10}\selectfont\itshape
$^{\sharp}$\,Sharif University of Technology
\vspace{1.2mm}\\
\fontsize{10}{10}\selectfont\rmfamily\itshape
$^{\star}$\,Queen Mary University of London

\vspace{1.2mm}\\
\fontsize{10}{10}\selectfont\rmfamily\itshape
$^{\dagger}$\,Nokia Bell Labs \& University of Oxford

\vspace{1.2mm}\\
\fontsize{10}{10}\selectfont\rmfamily\itshape

}
\begin{document}
\maketitle
\begin{abstract}

%

Deep neural networks are increasingly being used in a variety of machine learning applications applied to rich user data on the cloud. However, this approach introduces a number of privacy and efficiency challenges, as the cloud operator can perform secondary inferences on the available data. Recently, advances in edge processing have paved the way for more efficient, and private, data processing at the source for simple tasks and lighter models, though they remain a challenge for larger, and more complicated models. In this paper, we present a hybrid approach for breaking down large, complex deep models for cooperative, privacy-preserving analytics. We do this by breaking down the popular deep architectures and fine-tune them in a particular way. We then evaluate the privacy benefits of this approach based on the information exposed to the cloud service. We also asses the local inference cost of different layers on a modern handset for mobile applications. Our evaluations show that by using certain kind of fine-tuning and embedding techniques and at a small processing costs, we can greatly reduce the level of information available to unintended tasks applied to the data feature on the cloud, and hence achieving the desired tradeoff between privacy and performance.
\end{abstract}

%
%
%

\section{Introduction} 
\label{intro}

The increasing availability of connected devices such as smartphones and cameras have made them an essential and inseparable part of our daily lives. Majority of these devices collect forms of data and transfer it to the cloud in order to benefit from cloud-based data mining services like recommendation systems, targeted advertising, security surveillance, health monitoring and urban planning. Many of these applications are free, relying on information harvesting from their users' personal data. This practice has a number of privacy concerns and resource impacts for the users~\cite{narseoIMC,dontkill}. Preserving individuals' privacy, versus detailed data analytics, face a dichotomy in this space. Cloud-based machine learning algorithms can provide beneficial or interesting services (e.g., video editing tools or health apps), however, their reliance on excessive data collection form the users can have consequences which are unknown to the user (e.g., face recognition for targeted social advertising).

While complete data offloading to a cloud provider can have immediate or future potential privacy risks \cite{pournajaf2016,haris2014privacy}, techniques relying on performing complete analytics at the user end (on-premise solution), or encryption-based methods, also come with their own resource limitations and user experience penalties (see Section~\ref{sec:related} for detailed discussions). Apart from the resource considerations, an analytics service or an app provider might not be keen on sharing their valuable and highly tuned models. Hence, it is not always possible to assume local processing (e.g., a deep learning model on a smartphone) is a viable solution even if the task duration, memory and processing requirements are not important for the user, or tasks can be performed when the user is not actively using their device (e.g., while the device is being charged overnight).

In this paper, we focus on achieving a compromise between resource-hungry local analytics, versus privacy-invasive cloud-based services. We design and evaluate a hybrid architecture where the local device and the cloud system collaborate on completing the inference\footnote{In this paper, by inference we mean applying a pre-trained deep model on an input to obtain the output, which is different from statistical inference.} task. In this way, we can augment the local device to benefit from the cloud processing efficiency while addressing the privacy concerns. We concentrate on data mining applications where in order to get certain services from a provider, sending the data to the cloud is inevitable. As a specific exemplar of this general class of services, we consider image processing applications using deep learning. We address the challenge of performing certain approved image analytics in the cloud, without disclosing important information which could lead to other inferences such as identity leak via face recognition.

As an exemplar use case for this paper, we consider a case where we wish to enable specific inference tasks such as gender classification or emotion detection on face images, while protecting against a privacy-invasive task such as face recognition by a cloud operator having access to rich training data and pre-trained models (e.g., Google and Facebook). Convolutional Neural Networks (CNNs) are one of the most powerful instances of deep neural networks for doing image analysis \cite{Rich:2016:TBA:2896338.2897734, Druzhkov2016, wan2014deep}, and we use them to build accurate gender and emotion predictor models. We will fine-tune these models with our suggested architecture which brought us identity privacy, while still keep them accurate (As shown previously in~\cite{DBLP:journals/corr/OssiaSTRLH17}). We will perform our evaluations on smartphones, but it can also be extended to other devices with limited memory and processing capabilities e.g., Raspberry Pi and edge devices. A number of works \cite{lane2015can, lane2015zoe, kim2015compression} address the problem of using deep models on smartphones. However, using complex and accurate models in smartphones, requires significant processing and memory resources and our solution could highly improve their efficiency.

Our approach relies on optimizing the layer separation of pre-trained deep models. Primary layers are held on the user device and the secondary ones on the cloud. In this way, the inference task starts by applying the primary layers as the feature extractor on the user device, and continues by sending the resultant features to the cloud and, end by applying the secondary analyzing layers in cloud. We demonstrate that our proposed solution does not have the overhead of executing the whole deep model on the user device, while it will be favored by a cloud provider as the user does not have access to their complete model and part of the inference should be done on the cloud. We introduce a method to manipulate the extracted features (from the primary layers) in a way that irrelevant extra information can not leak, hence addressing the privacy challenges of cloud solution. To do this, we alter the training phase by applying Siamese network \cite{chopra2005} in a specific manner, and by employing a dimensionality reduction and noise addition mechanism for increased privacy.  


In Section~\ref{evaluation}, we use three methods to quantify the privacy guarantees of our approach. One is to use \textit{transfer learning} \cite{yosinski2014} which proves that face recognition is impractical by even using the state of the art models. The second approach is to use deep visualization techniques, which tries to reconstruct the input image by just using the extracted feature in the intermediate layer \cite{dosovitskiy2016}. At the end we introduce a new metric for privacy measurement which is an extension for optimal Bayes error. We also implement our model on smartphone and compare the fully on-premise solution with the hybrid solution presented in this paper.\footnote{Our codes are available at \url{https://github.com/aliosia/DeepPrivInf2017}}
 

Our main contributions in this paper include:
\begin{itemize}
	\item Proposing a learning framework for privacy-preserving analytics on a cloud system and embedding deep networks on it;
	\item Developing a new technique for training deep models based on the Siamese architecture, which enables privacy at the point of offloading to the cloud;
	\item Performing evaluation of this framework across two common deep models: VGG-16 and VGG-S and two applications: gender classification and emotion detection in a way that we preserve the privacy relating to face recognition.
\end{itemize}

%
%
%
%
%




\label{decomposition}
\section{Hybrid Framework}


\begin{figure}[t]
	\begin{center}
		\includegraphics[width=.9\columnwidth]{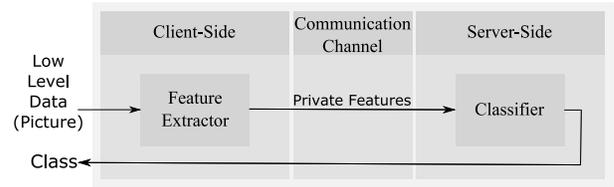} 	
	\end{center}
	\caption{Hybrid privacy-preserving framework.}
	\label{fig:framework}
\end{figure}

In this section, we present a hybrid framework for privacy preserving analytics. Suppose we want to utilize a cloud service to infer a \emph{primary} information of interest (e.g., gender, age or emotion on video footage or images), and at the same time, we ought to prevent the exposure of \emph{sensitive} information (e.g., identity) to the cloud provider. Hence, the data shared with the cloud service should possess two important properties: (i) inferring the primary information is possible; and (ii) deducing the sensitive information is not possible. The only solution to build this data, is to process the raw data on the client side and extract a rich feature with this properties. We can then transfer this feature to the cloud for further processing, without initial privacy concerns. Hence, we can consider a hybrid framework, in which the user and the service provider cooperate with each other. Figure~\ref{fig:framework} presents an overview of this framework. We can break down the analytics process into feature extraction and analyzation:

\begin{itemize}
	\item \textit{Feature Extractor}: This module takes the raw input data, process it, and outputs a rich feature vector which needs to keep the \emph{primary} information, while it should protect the \emph{sensitive} information. Usually, these two objectives are contradictory, i.e., decreasing the sensitive information causes a decrease in the primary information too. Additionally, due to limitations of client side processing, the feature extraction task needs to have minimal burden; consequently the designing the feature extractor is the most challenging task. 
	
	
	\item \textit{Analyzer}: This module takes the intermediate features, generated by the feature extractor, as its input, and analyzes it. In practice, this module can be any ordinary classifier and the privacy of intermediate features was ensured by the first module.
\end{itemize}

We also need a protocol between the service provider and user to establish this framework. Suppose the service provider knows about the primary (e.g. gender) and sensitive (e.g. identity) user information. Because of that, the feature extractor can be designed by the service provider and yielded to the client. This feature extractor is guaranteed to consider user's primary and sensitive information, simultaneously. Demonstrating that the primary information is kept in the features can be done by showing the efficiency of the Analyzer. The service provider should also define a verification method for the privacy preservation; different methods for doing this is discussed in Section~\ref{evaluation}. 


Our framework is generic and can be used for any privacy-preserving learning problem. In Section~\ref{DLembedding}, we explain how to embed feedforward neural networks in this framework.

\label{theory}
\newcommand\norm[1]{\left\lVert#1\right\rVert}

\section{Deep-Priv Embedding}
\label{DLembedding}

\begin{figure}[t]
	\centering
	\begin{subfigure}[b]{.9\columnwidth}

		\includegraphics[width=\columnwidth]{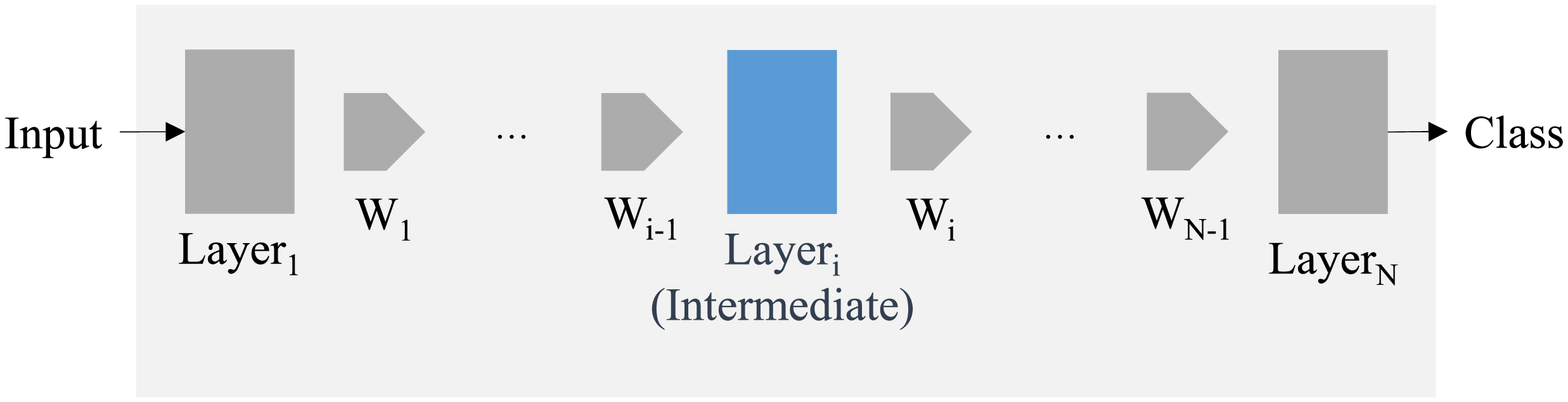}
		\caption{Training simple embedding.}
		\label{fig:simpletrain}
	    \vspace*{15pt}
	\end{subfigure}

	\begin{subfigure}[b]{.9\columnwidth}  

		\includegraphics[width=\columnwidth]{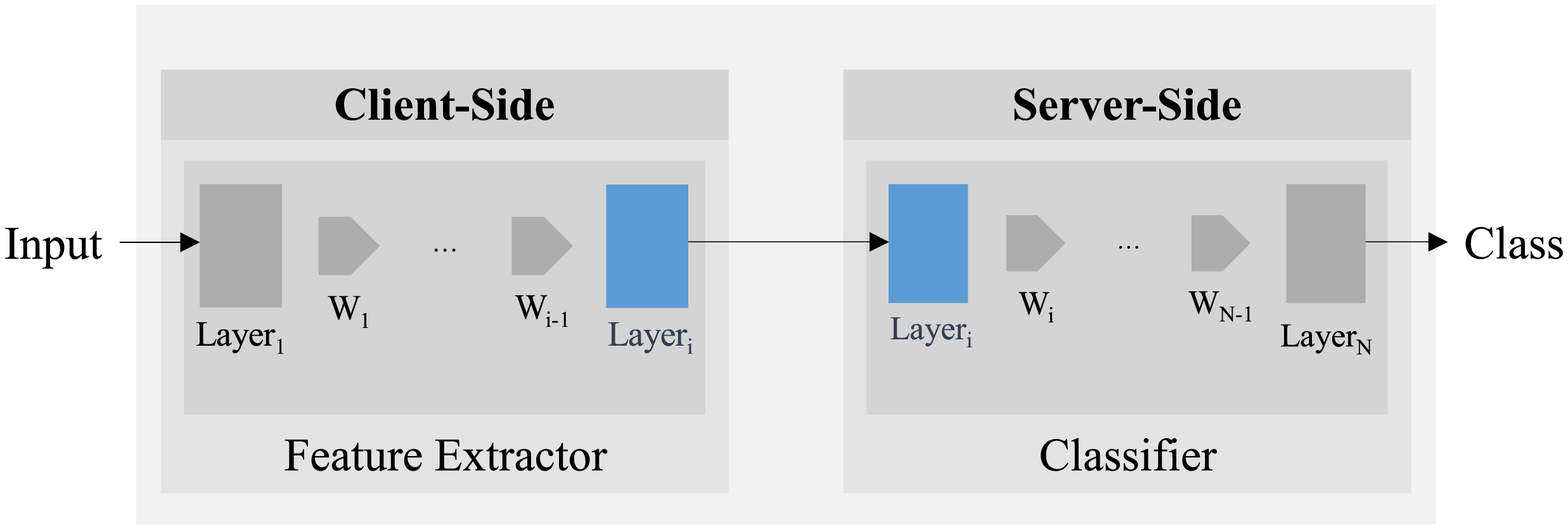}
		\caption{Using simple embedding. Intermediate feature is passed through communication channel.}
		\label{fig:simpletest}
	\end{subfigure}
	\vspace{5pt}
	\caption{Simple embedding of a deep network.}
	\label{fig:simple}
\end{figure}

Due to the increasing popularity of deep models in analytics applications, in this section we address how to embed an existing deep model, inferring primary information (e.g. predicting gender or emotion) in the proposed framework. Complex deep networks consist of many layers which can be embed in this framework, using a layer separation mechanism. First, we should choose the intermediate layer from a deep network and then we can store the layers before that on the client device as a feature extractor, and the layers after that in the cloud server as the classifier (see Figure~\ref{fig:framework}). Choosing the intermediate layer from higher layers of the network, intrinsically comes with privacy compromises. In~\cite{mahendran2015}, the authors reconstruct an original image from each layer and the accuracy of reconstruction decreases by using higher layers. As we go up through the deep network layers, the features get more specific to the primary information~\cite{yosinski2014} and irrelevant information (including sensitive information) will be gradually lost. Hence, by using the layer separation mechanism, we achieve two important objectives simultaneously: (i) we end up with the feature extractor easily, and (ii) we benefit from the intrinsic characteristics of deep models. This approach satisfies the initial criteria we set for our proposed framework. In this paper, we refer to this embedding as the \emph{simple embedding}. The training and test phase of this embedding can be seen in Figure~\ref{fig:simple}. 

Moreover, experiments show that the accuracy of primary classification does not decrease, when we reduce the dimension of the intermediate feature with Principle Component Analysis (PCA). This can improve privacy due to intrinsic characteristics of dimensionality reduction. We can also highly reduce the communication overhead between the client and server. We refer to this embedding (with PCA applied) as the \textit{reduced simple embedding}.
 
An important challenge with deep models in privacy applications, is that they learn invariant general features which are not specific to the target task \cite{bengio2012deep}. This characteristic of deep networks, adversely affects their privacy. The solution is to manipulate the intermediate feature and try to specialize it for the primary variable and make the sensitive variable unpredictable. One way to do this is to have a many to one mapping for the sensitive variable. This is the main idea behind \emph{k-anonymity}~\cite{sweeney2002}, assuming the identity is the sensitive variable. As an example, Suppose $k$ different male images are mapped to one point in the feature space. Having this feature, an attacker will have confusion between $k$ possible identities. We use the Siamese architecture~\cite{chopra2005} to accomplish this task, as much as possible. To the best of our knowledge, this is the first time that the Siamese architecture is used as a privacy preservation technique. Fine-tuning with Siamese architecture results in a feature space where objects with the same primary classes cluster in together. Due to this transformation, borders of the sensitive variable classes get faded, consequently classifying sensitive variable becomes harder or even impossible, while the primary information is not affected. We refer to this embedding as the \emph{Siamese embedding}, where Siamese fine-tuning is applied. In addition, we can reduce the dimensions of the intermediate feature without any deficiency; we refer to this embedding method as the \emph{reduced Siamese embedding}.

\begin{figure}[t]
	\centering
	\begin{subfigure}[b]{.9\columnwidth}    
		\includegraphics[width=\textwidth]{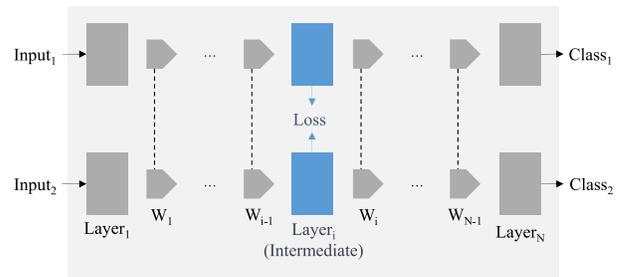}
		\caption{Training advanced embedding with Siamese architecture. Weights connected by dashed lines are equal.}
		\label{fig:advancedtrain}
		\vspace*{15pt}
	\end{subfigure}
	\begin{subfigure}[b]{.9\columnwidth}   
		\includegraphics[width=\textwidth]{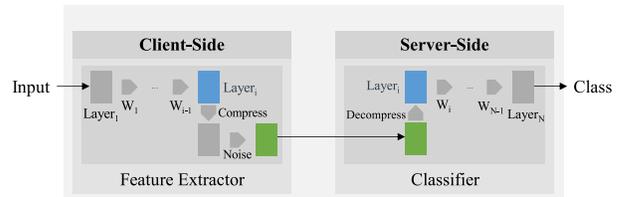}
		\caption{Using advanced embedding (with PCA projection and noise addition in client side and reconstruction and analyzing in server side).}
		\label{fig:advancedtest}
	\end{subfigure}
	\vspace{5pt}
	\caption{Advanced embedding of a deep network}
	\label{fig:advanced}
\end{figure}


Another method which increases the client privacy and inference uncertainty of unauthorized tasks is noise addition. A service provider can determine a noise addition strategy for its clients in order to increase the uncertainty of other undesired tasks. We refer to \textit{noisy embedding} whenever we use noise addition within the feature extractor. We also refer to the noisy reduced Siamese embedding as the \textit{advanced embedding}. In order to see the effect of Siamese fine-tuning, dimensionality reduction and noise addition, advanced embedding is shown in Figure~\ref{fig:advanced}. Hence in the feature extractor module of advanced embedding, the following steps should be taken:
\begin{itemize}
	\item{Applying primary layers.}
	\item{Reducing the dimensionality.}
	\item{Adding noise.}
\end{itemize}
The analyzer module should also do these steps:
\begin{itemize}
	\item{Reconstructing the feature vector.}
	\item{Applying remaining layers.}
\end{itemize}

In what follows, we discuss our Siamese fine-tuning, dimensionality reduction and noise addition strategy in details.



\subsection{Siamese Fine-tunning}

\begin{figure}[t]
	\centering
	\begin{subfigure}[b]{0.48\columnwidth}
		\includegraphics[width=.9\columnwidth, height=4cm]{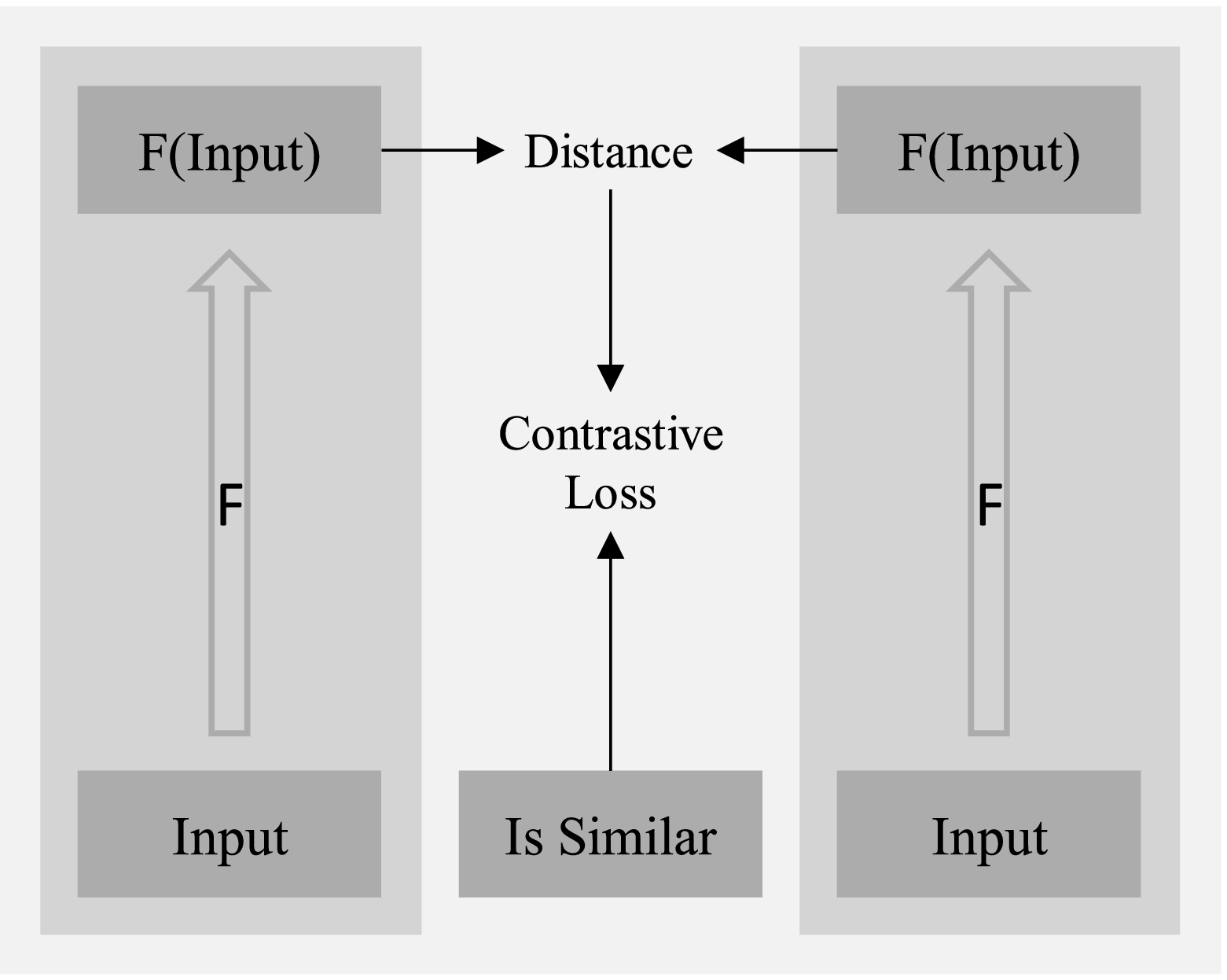}
		\vspace{5pt}
		\caption{Traditional Siamese arch.}
		\label{fig:siamese}
	\end{subfigure}%
	\quad
	\begin{subfigure}[b]{0.48\columnwidth}
		\includegraphics[width=.9\columnwidth, height=4cm]{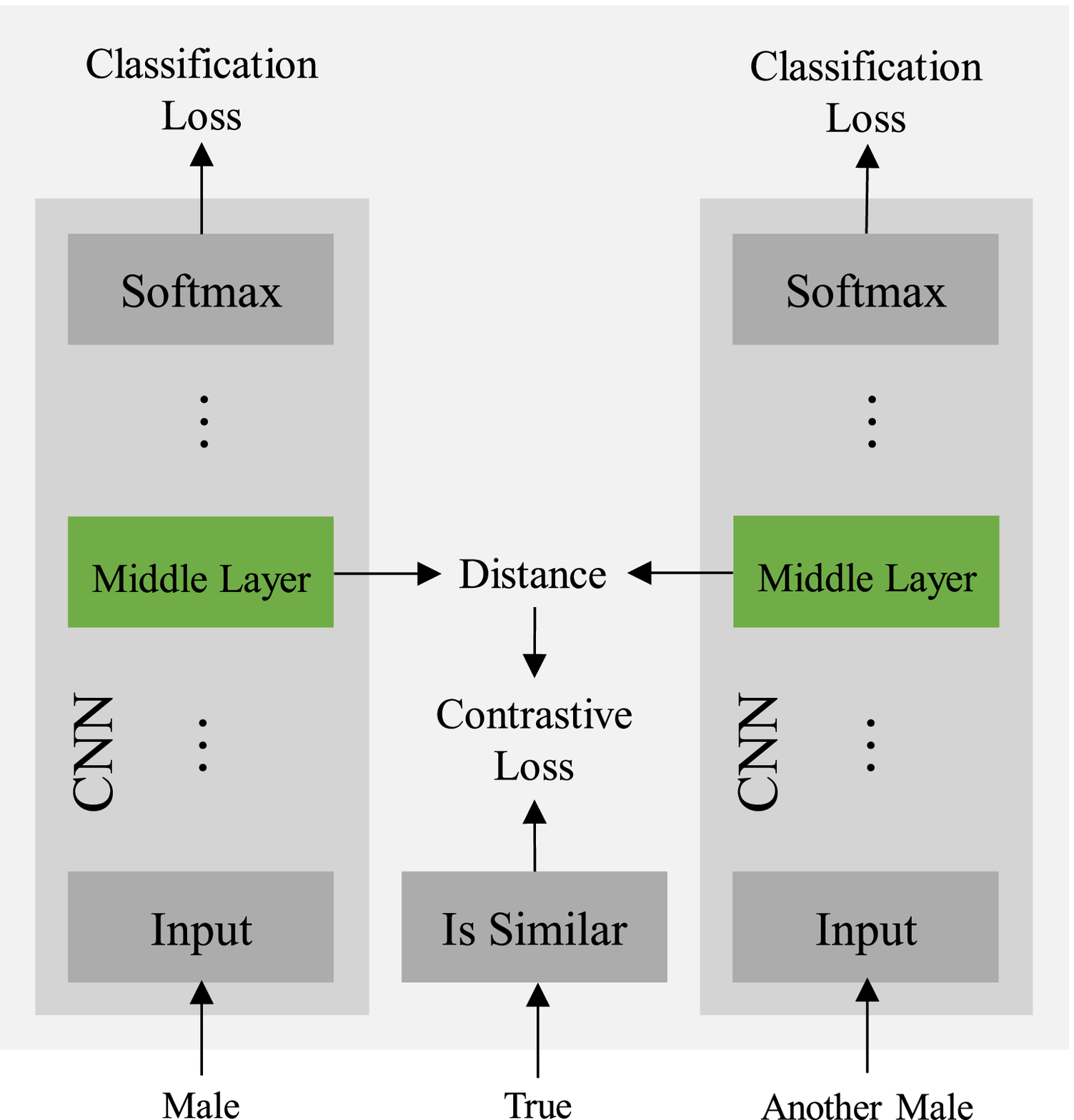}
		\vspace{5pt}
		\caption{Siamese arch. for privacy}
		\label{fig:siam_priv}
	\end{subfigure}%
	\vspace{10pt}
	\caption{Siamese architecture usage}
	\label{fig:siamese_model}
\end{figure}

The Siamese architecture has previously been used in verification applications~\cite{chopra2005}. It provides us with a feature space, where similarity between the data points is defined by their euclidean distance. The main idea of training with Siamese architecture is forcing the representations of two similar points to become near each other, and the representations of two dissimilar points become far. In order to do this, our training dataset should consists of pairs of similar and dissimilar points. For a pair of points, one function is applied to both of them and their value distance is computed. A contrastive loss function should be defined in a way that making this distance maximize for two dissimilar points and being minimized for two similar points. An appropriate such loss function is defined in~\cite{hadsell2006} and we use it in our application:
\begin{align}
L(f_1,f_2) = \begin{cases}
\norm{f_1-f_2}_2^2 & \quad \text{\small similar}\\
\max(0,margin - \norm{f_1-f_2}_2)^2 & \quad \text{\small dissimilar}
\end{cases}
\end{align}
where $f_1$ and $f_2$ are the mappings of data points. The traditional Siamese architecture is presented in Figure~\ref{fig:siamese}. As an example, consider the face verification application. We want to determine whether two images belong to the same person or not. We should prepare a dataset consists of pairs of face images, some of them are similar and belong to the one person and some are not. Then by using a convolutional neural network as a feature extractor and imposing a contrastive loss function, we can train a similarity metric between face images.  

How can we use this architecture in privacy-preserving analytic application? as we said before, we have a pre-trained deep network which predict primary variable. How can we make one of its intermediate layer private to sensitive information? Our proposed solution is to define a contrastive loss on the intermediate layer and build a multi-objective optimization problem, which tries to increase both privacy of sensitive variable and accuracy of primary variable prediction. We just need to properly define similarity between pairs of input data. 

Assuming the sensitive variable is identity and the primary variable is gender (or emotion), we can define two face image with the same gender (or emotion) and different identity as similar. Using this way, we try to map different identities from the same gender to the same point which brought us privacy. The architecture of this network is presented in Figure~\ref{fig:siam_priv}.

\subsection{Dimensionality Reduction}



In order to increase privacy and decrease communication cost, the service provider could reduce the dimensionality of the intermediate feature by applying PCA or auto-encoder. In this way, the last layer of the feature extractor and the first layer of the analyzer should be a dense reduction and reconstruction matrices, respectively.  As we will show in Section~\ref{experiments}, this procedure does not affect significantly on the primary task accuracy.  


\subsection{Noise Addition}

\begin{figure}[t]	
	\centering
	\includegraphics[width=.9\columnwidth]{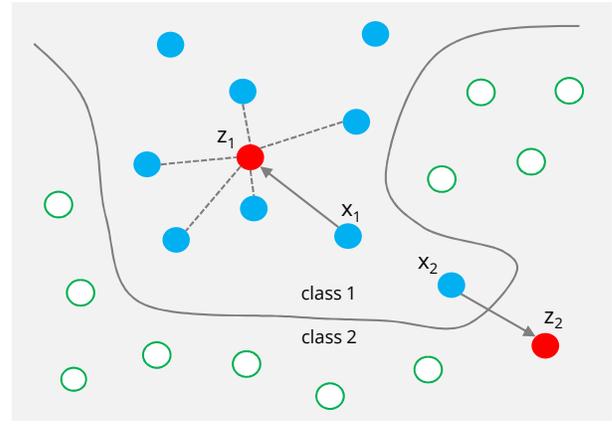} 	
	
	\caption{The effect of noise addition on the two dimensional feature space. Blue and green points show the objects of the first and second primary class, respectively. $z_1$ and $z_2$ have been obtained by adding noise to $x_1$ and $x_2$, respectively.}
	\label{fig:noiseEffect}
\end{figure}

Succeeding the dimensionality reduction, we can add a multidimensional noise to the feature vector, to further increase the privacy. Siamese fine-tuning tries to map some objects with different sensitive classes (e.g. identity) to the same point, while in practice, these points may have small distances from each other. We can highly increase the uncertainty about the sensitive variable by adding a random noise to it (see $x_1$ and $z_1$ in Fig.~\ref{fig:noiseEffect}). However, this task may decrease the accuracy of the primary variable prediction (see $x_2$ and $z_2$ in Fig.~\ref{fig:noiseEffect}). Thus, we face a trade-off between privacy and accuracy, while increasing the amount of noise. Siamese fine-tuning makes this trade-off significantly better than the noisy reduced simple embedding, without fine-tuning in the way discussed. The reason is that while doing the Siamese fine-tuning, the intra-class and inter-class variance is decreased and increased for the primary variable, respectively. Experiments in Section~\ref{experiments} confirm this conclusion, by testing different variance for a multi-dimensional symmetric Gaussian noise and observing the trade-off.

\section{Privacy Measurement}
\label{evaluation}

In this section, we introduce three different ways to evaluate the privacy of the feature extractor:
\begin{enumerate}
	\item Transfer Learning approach~\cite{yosinski2014} can be used to determine the degree of generality and specificity of the extracted features. 
	\item Deep visualization~\cite{dosovitskiy2016} evaluate the capability of reconstructing back the input image. 
	\item Probabilistic modelling of the sensitive variable can be also helpful for defining a metric for privacy. 
\end{enumerate}
In the following, we discuss about each of these methods.


\subsection{Transfer Learning}
\label{transfer}

\begin{figure}[t]
	\centering
	\begin{subfigure}[b]{.9\columnwidth}    
		\includegraphics[width=\textwidth]{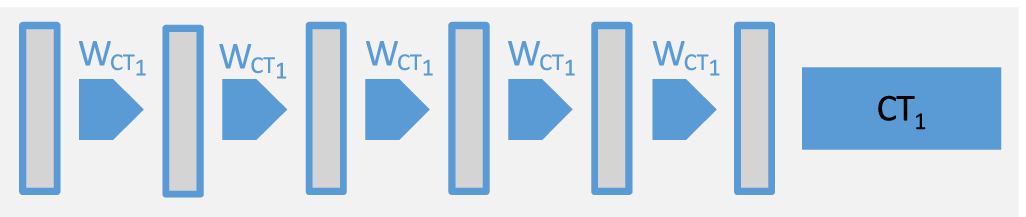}
		\caption{Trained network for primary classification ($N_1$)}
		\label{fig:transfer1}
		\vspace*{15pt}
	\end{subfigure}
	\begin{subfigure}[b]{.9\columnwidth}    
		\includegraphics[width=\textwidth]{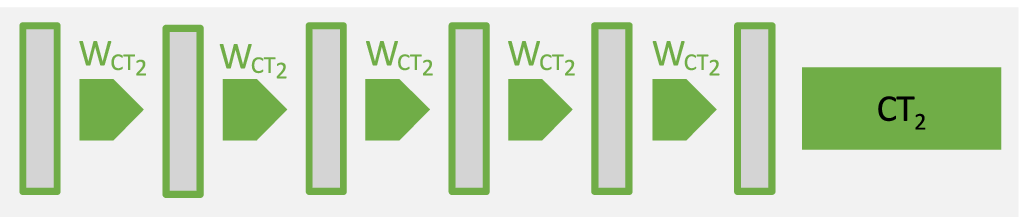}
		\caption{Network for sensitive variable inference ($N_2$)}
		\label{fig:transfer2}
		\vspace*{15pt}
	\end{subfigure}
	\begin{subfigure}[b]{.9\columnwidth}   
		\includegraphics[width=\textwidth]{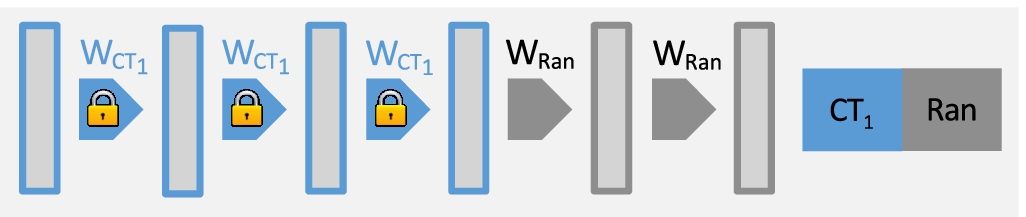}
		\caption{Primary weight are copied from $N_1$ and frozen. The other layers have random weights.}
		\label{fig:transfer3}
		\vspace*{15pt}
	\end{subfigure}
	\begin{subfigure}[b]{.9\columnwidth}   
		\includegraphics[width=\textwidth]{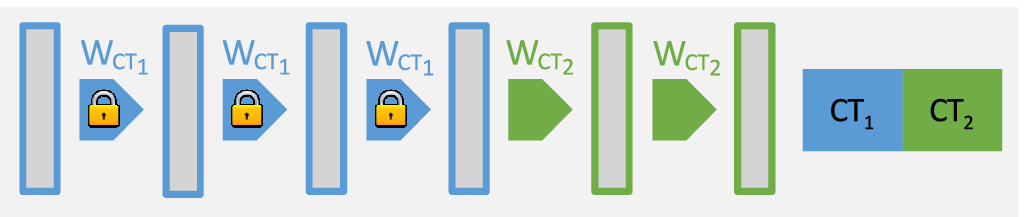}
		\caption{Trained network on sensitive variable inference with transfer learning}
		\label{fig:transfer4}
	\end{subfigure}
	\vspace{5pt}
	\caption{Transfer Learning procedure.}
	\label{fig:transfer}
\end{figure}

We can measure the amount of specificity of the extracted feature to the primary task by using transfer learning~\cite{yosinski2014}. Suppose we have a trained network $N_1$ for primary classification (Figure~\ref{fig:transfer1}). We build and train network $N_2$ for sensitive variable inference (Figure~\ref{fig:transfer2}) with the following procedure:
\begin{itemize}
	\item Copy weights from the first $i$ layers of $N_1$ to the first $i$ layers of $N_2$;
	\item Initialize the reminding layers of $N_2$ randomly (Figure~\ref{fig:transfer3});
	\item Freeze the first $i$ layers of $N_2$ (do not update their weights);
	\item Train $N_2$ for sensitive variable inference (Figure~\ref{fig:transfer4}).
\end{itemize}

After the training procedure, the accuracy obtained for sensitive variable prediction is directly related to the degree of specificity or generality of the extracted feature from $i$'th layer. As we get lower general accuracy for sensitive variable prediction, the feature is more specific to primary task.

\subsection{Deep Visualization}

Visualization is a method for understanding the deep networks. In this paper, we used an auto-encoder objective visualization technique \cite{dosovitskiy2016}  in order to measure the amount of sensitive information in the intermediate feature of the network, which is trained for primary variable inference. In \cite{dosovitskiy2016}, a decoder is designed on the data representation of each layer, in order to reconstruct the original input image based on the learned representation. So, we can analyze the preserved sensitive information in each layer, via comparing the reconstructed images with the original input image.

\subsection{Privacy Metric}
\label{validation}

Suppose we have an estimate for the posterior distribution of the sensitive variable (e.g. identity), given the extracted feature vector. It can be obtained by using a simple instance-based model like kernel density estimation or a complex neural network. How can we measure the amount of information existed in this distribution? Conditional entropy and classification Bayes error could be different options for information measurement; but, here we are going to introduce a more intuitive method to measure privacy, which is the extension of Bayes error. In order to get more accurate results, here we assume that we use both dimensionality reduction and noise addition.

Suppose we have a dataset and we want to measure the privacy level of the feature extractor. We can get all the intermediate features and apply noise to them. Having all these features $(\{f_i\})$ and a fixed noisy data point such as $z$, we can calculate the conditional likelihood of each sensitive classes. In order to do this, we can estimate $P(z | c_i)$ in this way:

\begin{equation}
\begin{split}
P(z | c_i) & = \int_f P(z,f | c_i) df \\
& = \int_f P(z | f,c_i)P(f | c_i) df
\end{split}
\end{equation}
Conditioned on $f$, $c_i$ is independent of $z$, so we have:
\begin{equation}
\begin{split}
P(z | c_i) & = \int_f P(z | f)P(f | c_i) df\\ 
& = E_{f \sim P(f|c_i)} [P(z | f)]
\end{split}
\end{equation}
Assuming $F_i=\{f_1, f_2, ... , f_{N_i}\}$ is the set of points from sensitive class $c_i$ in our dataset, we can estimate the above expected value with sample mean; so we can estimate $P(z | c_i)$ with:
\begin{equation}
\widehat{P(z|c_i)} = \frac{1}{N_{c_i}} \sum_{f_j \in F_i}P(z | f_j)
\end{equation}

In this way, we can compute the relative likelihood of each class given a noisy data point. As we know the correct class of that point, we can determine the number of classes with a higher probability than the correct class. Hence, we can define the rank of the likelihood of the right class, as the privacy of that noisy point. We want this measure to have a normalized value between 0 and 1, so we divide it by $T$, the number of sensitive classes:
$$Privacy(z)=\frac{Rank(class(z))}{T}$$

Now, having intermediate features of $N$ samples (with $N$ noisy points generated by them), we can estimate the privacy of the transmitted data by:
$$Privacy_{\ total} = \sum_{i=1}^N \frac{Privacy(z_i)}{N}$$

We can define this as a measure for quantifying privacy. In the next sections, we simply refer to this metric as \textit{Privacy}. With this measure, we can calculate how much privacy is preserved and also validate the privacy of the transmitted data.

\label{applications}
\section{Applications}
 
In this section, we introduce gender classification and emotion detection as two exemplar primary tasks and consider the face identity as the sensitive information; so we use the VGG-16 face recognition model~\cite{Parkhi15} as the adversary, trying to infer the sensitive variable. We evaluate transfer learning approach by using the IMDB dataset used in~\cite{Parkhi15} which contains near 2 million images from 2,622 highly-ranked celebrities on the IMDB website. We randomly select 100 celebrities and divide their images to training and test sets to evaluate our face recognition model. 


\subsection{Gender Classification}

\begin{figure}[t]
	\begin{center}
		\includegraphics[width=.9\columnwidth]{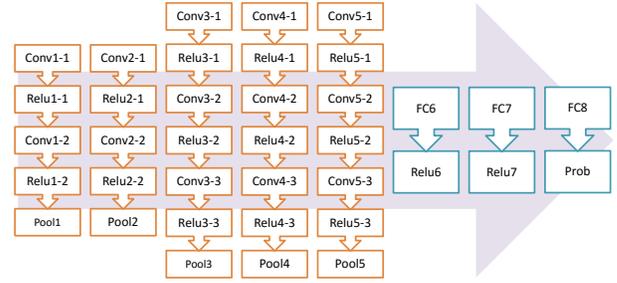} 	
	\end{center}
	\caption{16 layer VGG-16 architecture \cite{simonyan2014very}}
	\label{fig:vgg-16}
\end{figure}

In the problem of gender classification, the  goal is to classify an individuals' image to Male or Female. This has various applications in different systems such as human-computer interaction, surveillance and targeted advertising systems~\cite{ng2012}. Some techniques use face image as the input to the classifier, while others use the whole body image or a silhouette. In this paper, we use cropped face images for the gender classification task. Recently, deep convolutional neural networks have been used for this problem~\cite{rachaud2015, rothe2015, levi2015age}. In this work we use the model proposed in~\cite{rothe2015} with 94\% accuracy, based on VGG-16 architecture, the popular 16-layer deep model for image classification~\cite{simonyan2014very} (see Figure~\ref{fig:vgg-16}).

Rothe~\etal~\cite{rothe2015} prepared a huge dataset, named IMDB-Wiki, which is useful for age and gender estimation. We use the Wiki part of this dataset which contains 62,359 images to fine-tune our models. We use 45,000 images as training data and the rest as test data. We evaluate our privacy measurement technique on this dataset. We also use Labeled Face in the Wild (LFW) dataset~\cite{LFWTech} to compare our gender classification model with others. This is an unconstrained face database containing 13,233 images of 5,749 individuals which is very popular for evaluating face verification and gender classification models. 

\subsection{Emotion Detection}

\begin{figure}[t]
	\begin{center}
		\includegraphics[width=.9\columnwidth]{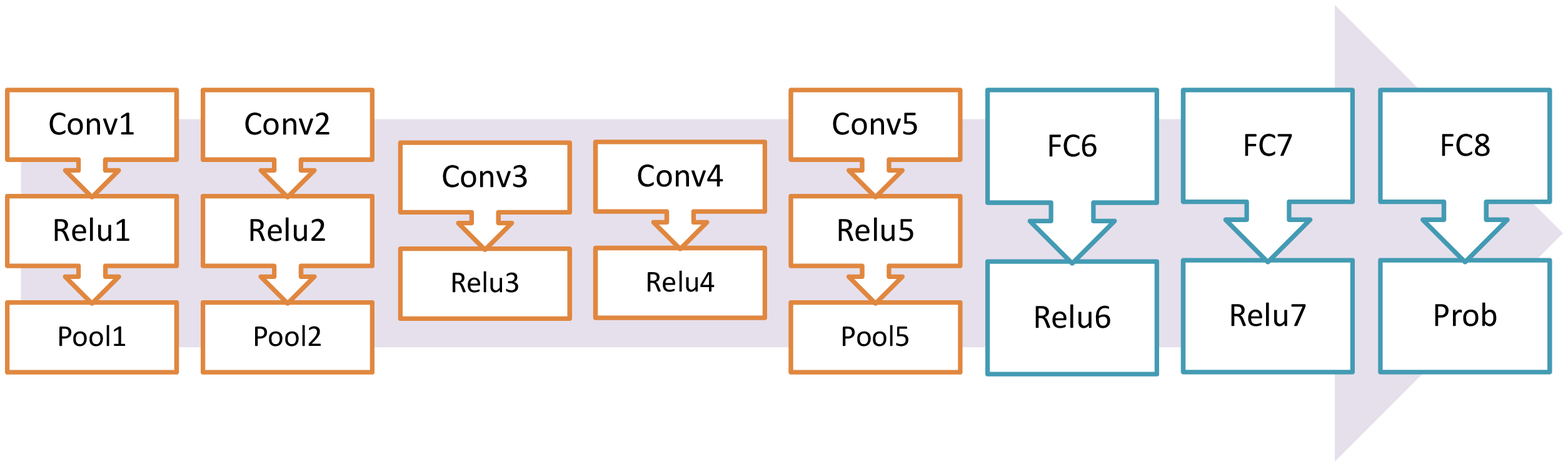} 	
	\end{center}
	\caption{8 layer VGG-S architecture \cite{chatfield14}}
	\label{fig:vgg-s}
\end{figure}

Emotion detection from facial expression is becoming exceedingly important for social media analysis tasks. In this problem, emotions are classified based on the individuals' facial expressions on images. Recently, deep learning has been demonstrated to be effective in solving this problem~\cite{levi2015emotion, mollahosseini2016}. Different deep models are proposed and compared in \cite{levi2015emotion}. We choose the \textit{VGG-S\_RGB} model which is based on VGG-S architecture~\cite{chatfield14} (see Figure~\ref{fig:vgg-s}). The accuracy of doing emotion detection by using this model is 39.5\% on SFEW-2 dataset. 
Static Facial Expression in the Wild (SFEW) is an emotion detection benchmark~\cite{dhall2011static}. We use the latest version~\cite{dhall2015video} which consists of face images in seven emotional classes. This dataset contains 891 and 431 images for training and validation respectively. 

\section{Experiments}
\label{experiments}

In this section we evaluate and analyze the accuracy and privacy of different embeddings with different intermediate layers, by using our proposed privacy measurement tools: transfer learning, visualization and privacy metric. Although all of these embedding preserve privacy, applying Siamese fine-tuning is more efficient in a way that it increase privacy considerably, whereas it does not decrease the accuracy of primary task. In addition, we show how dimensionality reduction has positive effects on privacy. Finally, we evaluate our hybrid framework on mobile phone and discuss its advantages regarding to other solutions.

\subsection{Privacy of Gender Classification}

In this part, we apply transfer learning, privacy metric and visualization technique on different intermediate layers of gender classification and face recognition models, in order to show the privacy of our framework. We use the VGG-16 model proposed at \cite{rothe2015} in the simple embedding and fine-tune it with the proposed privacy architecture (Figure~\ref{fig:siam_priv}) to use it in Siamese embedding. 
 To create the reduced simple and Siamese embeddings, we apply PCA on the intermediate features of simple and Siamese embeddings, respectively. We choose 4, 6 and 8 as the PCA dimension for Conv5\_3, Conv5\_2 and Conv5\_1 respectively. 

\begin{figure}[t]
	\begin{tikzpicture}
	\begin{axis}[
	ybar,
	label style={font=\small},
	ticklabel style = {font=\small},
	enlarge x limits=0.2,
	legend style = { font=\small, column sep = 1pt, legend columns = 1, at={(.99,.99)}, anchor=north east},
	y label style={font=\small, at={(axis description cs:.1,.5)},anchor=south},
	ylabel={Face Rec. accuracy (\%)},
	symbolic x coords={Conv5-1,Conv5-2,Conv5-3},
	xtick=data, ymin=0,
	nodes near coords,
	every node near coord/.append style={font=\tiny},
	nodes near coords align={vertical},
	]
	\addplot[fill=darkgreen!70!white,postaction={pattern=north east lines}] coordinates {(Conv5-1,29) (Conv5-2,24) (Conv5-3,15)};
	\addplot[fill=darkblue!70!white,postaction={pattern=north west lines}] coordinates {(Conv5-1,5.6) (Conv5-2,4.9) (Conv5-3,3.6)};
	\addplot [fill=rosso!70!white,postaction={pattern=horizontal lines}] coordinates {(Conv5-1,4.3) (Conv5-2,3) (Conv5-3,2.3)};
	\addplot [fill=giallo!70!white] coordinates {(Conv5-1,2.0) (Conv5-2,2.8) (Conv5-3,2.6)};
	
	\legend{simple, reduced simple, Siamese, reduced Siamese}
	\end{axis}
	\end{tikzpicture}
	\caption{Gender Classification. Comparison of simple, reduced simple, Siamese and reduced Siamese embedding on different intermediate layers, while doing transfer learning. 
	}
	\label{fig:FRGC}
\end{figure}
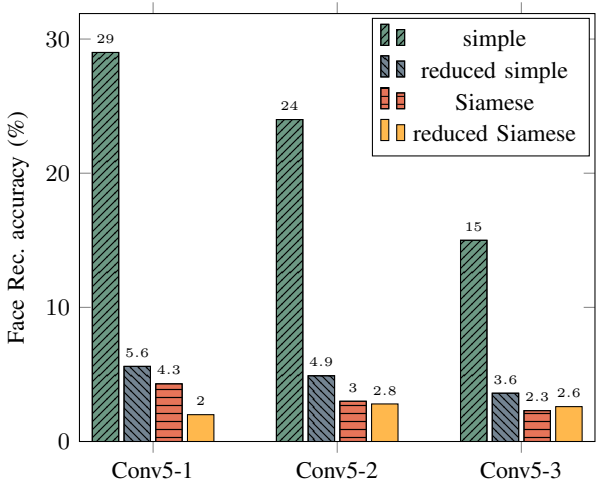
\subsubsection{Transfer learning}
The result of transfer learning for different embeddings on different intermediate layers are presented in Figure~\ref{fig:FRGC}. Overall, applying (reduced) simple or Siamese embedding results in a considerable decrease in the accuracy of face recognition from Conv5\_1 to Conv5\_3. The reason of this trend is that as we go up through the layers, the features of each layer will be more specific to the gender classification (primary task). That is to say, the features of each layer do not have information related to identity (sensitive information) as much as even its previous layer. In addition, for all of the layers, face recognition accuracy of Siamese embedding is by far less than the accuracy of simple embedding. This result has route in training of Siamese embedding with privacy architecture which causes a dramatic drop in the accuracy. As it is shown in Figure~\ref{fig:FRGC}, when Conv5\_3 is chosen as the intermediate layer in Siamese embedding, the accuracy of face recognition is 2.3\%, just ahead of random accuracy. Another interesting point of this figure is the effect of dimensionality reduction on the accuracy of face recognition. The reduced simple and Siamese embeddings has lower face recognition accuracy than simple and Siamese embedding, respectively.

\begin{table}[t]
	\caption{Accuracy of Gender Classification.}
	\begin{center}
		\begin{tabular}[width=\columnwidth]{ |c|c|c|c| }
			\hline
			\multicolumn{4}{|c|}{Accuracy on LFW} \\
			\hline
			& Conv5-1 & Conv5-2 & Conv5-3  \\
			\hline
			simple  & 94\% & 94\%  &  94\%  \\ 
			reduced simple& 89.7\%  &  87\% & 94\% \\
			Siamese & 92.7\% & 92.7\%  &  93.5\% \\
			reduced Siamese& 91.3\% &  92.9\% &  93.3\% \\
			\hline
		\end{tabular}
	\end{center}
	\label{tab:gcpca}
\end{table}

In order to assess the way these changes adversely affect accuracy of desired task which is gender classification, we report different embeddings accuracies in table~\ref{tab:gcpca}.  The results of table~\ref{tab:gcpca} convey two important messages. First, as the gender classification accuracy of Siamese and simple embedding are approximately the same, applying Siamese idea does not decrease accuracy of desired task. The other important result is that Siamese embedding is more robust to PCA than the simple embedding. In other words, gender classification accuracy of reduced Siamese embedding is close to Siamese embedding, whereas dimensionality reduction damage the accuracy of simple embedding. Figure~\ref{fig:FRGC} and table~\ref{tab:gcpca} show that applying the Siamese network and dimensionality reduction results in preserving privacy while gender classification accuracy does not decrease dramatically.

\begin{figure}[t]
	\centering
	\begin{subfigure}{0.48\columnwidth}
		\begin{tikzpicture}
		\begin{axis}
		[legend style = { font=\tiny, column sep = 1pt, legend columns = 1, anchor=north east,  nodes={scale=0.8}, legend image post style={scale=0.5}},
		width=1.15\columnwidth,
		ymin=84,ymax=96,
		xmin=0,xmax=40,
		grid = both,
		ytick={84,87,...,96},
		xlabel=Face Rec. Privacy (\%),
		y label style={font=\tiny, at={(axis description cs:.2,.5)},anchor=south},
		x label style={font=\tiny, at={(axis description cs:.5,-.05)},anchor=south},
		ylabel=Gender Class. Accuracy (\%),
		ticklabel style = {font=\tiny},
		y tick label style={
			/pgf/number format/.cd,
			fixed,
			fixed zerofill,
			precision=0,
			/tikz/.cd
		},
	    legend image post style={scale=0.5},
		]
		\addplot[color=blue, thick, densely dashed] table{simple_gender_conv53_rank.txt}; \addlegendentry{noisy reduced simple}
		\addplot[color=awesome, thick] table{siamese_gender_conv53_rank.txt}; \addlegendentry{advanced}
		\end{axis}
		\end{tikzpicture}
		\caption{\small{Comparison of presence or absence of Siamese fine-tunning on layer Conv5-3.}}
		\label{fig:gender_presence}
	\end{subfigure}%
	\quad
	\begin{subfigure}{0.48\columnwidth}
		\begin{tikzpicture}
		\begin{axis}
		[legend style = { font=\tiny, column sep = 1pt, legend columns = 2, anchor=north east, nodes={scale=0.8}, legend image post style={scale=0.5} },
		width=1.15\columnwidth,
		ymin=84,ymax=96,
		xmin=0,xmax=40,
		grid = both,
		ytick={84,87,...,96},
		xlabel= Face Rec. Privacy (\%),
		x label style={font=\tiny, at={(axis description cs:.5,-.05)},anchor=south},
		label style={font=\tiny},
		ticklabel style = {font=\tiny},
		]
		\addplot[color=viola, thick, densely dashed] table{siamese_gender_pool5_rank.txt}; \addlegendentry{pool5}
		\addplot[color=awesome, thick] table{siamese_gender_conv53_rank.txt}; \addlegendentry{conv5\_3}
		\addplot[color=blu, thick, densely dotted] table{siamese_gender_conv52_rank.txt}; \addlegendentry{conv5\_2}
		\addplot[color=darkgreen, thick, dashed] table{siamese_gender_conv51_rank.txt}; \addlegendentry{conv5\_1}
		\end{axis}
		\end{tikzpicture}
		\caption{\small{Comparison of different layers. Higher layers achieves better trade-off.}}
		\label{fig:gender_layers}
	\end{subfigure}%
	\caption{Accuracy-Privacy trade-off for gender classification using VGG-16 architecture.}
	\label{fig:gc_accpriv}

\end{figure}
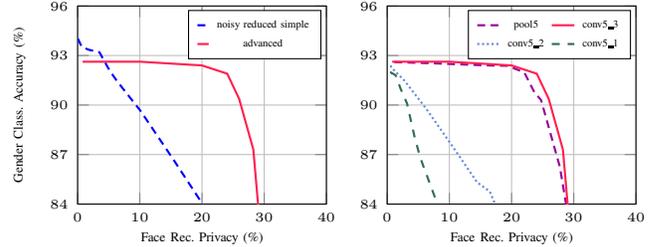

\subsubsection{Privacy metric}

In order to validate the feature extractor, we use the rank measure proposed in Section~\ref{validation}. By increasing the symmetric Gaussian noise variance, we get more privacy and less accuracy. In fact privacy and accuracy can be considered as two competing constraints in which increasing the identity privacy causes a decrease in accuracy of gender classification. We show this trade-off in Figure~\ref{fig:gender_presence}, where we can see the superiority of  the advance embedding (noisy reduced Siamese) over noisy reduced simple embedding. Figure~\ref{fig:gender_presence} is an evidence that by increasing privacy, gender classification accuracy decreases more slowly in advanced embedding than other embeddings. This makes the advanced embedding the ideal choice as we have better privacy on a fixed accuracy level. Another interesting experiment shows that choosing intermediate layers from higher ones, gives us better privacy for a fixed accuracy. This trend is shown in Figure~\ref{fig:gender_layers}, where the accuracy-privacy curve is upper for higher layers than lower ones and for a fixed accuracy, higher layer gives us more privacy. This validates our results of transfer learning in a way that choosing intermediate layers which are closer to the end of the network results in having a lower face recognition accuracy.

%

\subsubsection{Visualization}
\begin{figure*}[t]
	\centering
	\includegraphics[width = \textwidth]{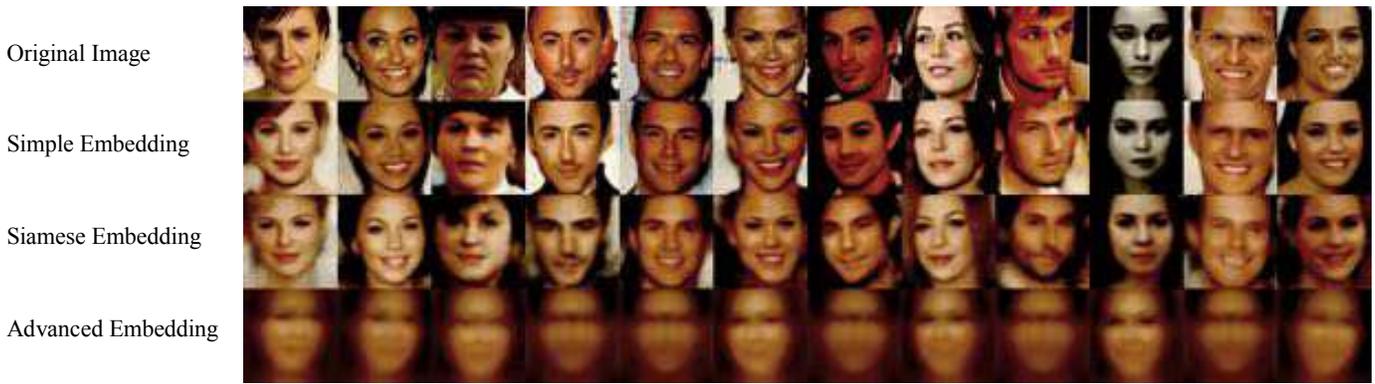} 	

	\caption{The first row shows the original images and the others show the reconstructed ones from intermediate representations. In all reconstructed images, the gender of the individuals is recognized to be the same as the originals. In addition, From simple to advanced embedding, the identity of the individuals is increasingly removed, illustrating that the \emph{advanced embedding} has the best privacy preservation performance.}
	\label{fig:visualization}
\end{figure*}

Deep visualization can brought us a good intuition about identity preservation of each layer. We fed the the intermediate layers of gender classification model as the input of Alexnet decoder \cite{dosovitskiy2016} to reconstruct the original inputs.  The reconstructed images leads to visually figure out the amount of identity information in the intermediate feature of gender classification model. These images are illustrated in Figure~\ref{fig:visualization} for different methods. It can be observed that the genders of all images in the simple and Siamese embeddings remain the same as the original ones. This is also the case for the advanced embedding, although it is harder to distinguish it from the reconstructed images. The original images are almost restored in the simple embedding. Therefore, just separating layers of a deep network can not assure acceptable privacy preservation performance. Siamese embedding performs better than the simple embedding by distorting the identity due to intrinsic characteristics of the face. Finally, the Advanced Embedding provides the best results, because the decoder was not trainable and nothing can be deduced from images, including the person's identity.

\subsection{Privacy of Emotion Detection}

We also evaluate our framework on emotion detection task. We use the VGG-S\_RGB pre-trained network of \cite{levi2015emotion} in the simple embedding. We fine tune their model with privacy architecture (Figure~\ref{fig:siam_priv}) on the training part of SFEW-2 dataset and get the Siamese embedding. As VGG-S has smaller structure in comparison with VGG-16 (8 layer vs. 16 layer), we just evaluate our embedding on one intermediate layer which is the fifth convolutional layer (Conv5). We choose 10 as the PCA dimension and get reduced simple and Siamese embedding.

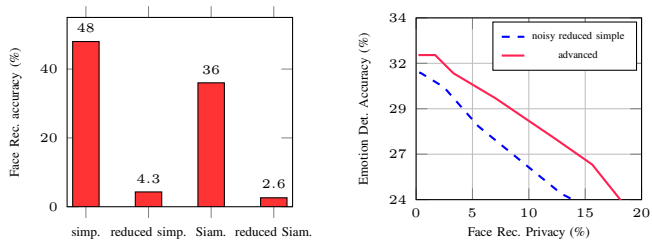
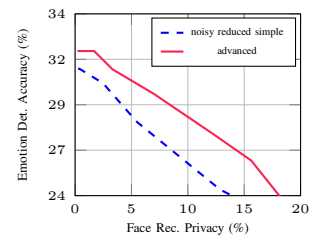
\begin{figure}[t]
	\centering
	\begin{subfigure}[b]{.47\columnwidth}
		\begin{tikzpicture}
		\begin{axis}[
		ybar,
		label style={font=\tiny},
		ticklabel style = {font=\tiny},
		width=1.10\textwidth,
		height= 4.1cm,
		ymax=55,
		legend style = { font=\tiny, column sep = 1pt, legend columns = -1, anchor=north east, nodes={scale=0.8}, legend image post style={scale=0.5}},
		ylabel={Face Rec. accuracy (\%)},
		y label style={font=\tiny, at={(axis description cs:.25,.5)},anchor=south},
		x label style={font=\tiny, at={(axis description cs:.5,-.05)},anchor=south},
		symbolic x coords={simp.,reduced simp., Siam., reduced Siam.},
		xtick=data, ymin=0,
		nodes near coords,
		every node near coord/.append style={font=\tiny},
		nodes near coords align={vertical},
		]
		\addplot [fill=red!80!white] coordinates {(simp.,48) (reduced simp.,4.3) (Siam.,36) (reduced Siam.,2.6)};
		\end{axis}
		\end{tikzpicture}
		\caption{Comparison of transfer learning results for different models.}
		\label{fig:FRED}
	\end{subfigure}
	\quad
	\begin{subfigure}[b]{.47\columnwidth}
		\begin{tikzpicture}
		\begin{axis}
		[legend style = { font=\tiny, column sep = 1pt, legend columns = 1, anchor=north east, nodes={scale=0.8}, legend image post style={scale=0.5}},
		width=1.10\textwidth,
		height= 4cm,
		ymin=24,ymax=34,
		ytick={24,26.5,...,34},
		xmin=0,xmax=20,
		xmajorgrids,
		ymajorgrids,
		xlabel=Face Rec. Privacy (\%),
		y label style={font=\tiny, at={(axis description cs:.25,.5)},anchor=south},
		x label style={font=\tiny, at={(axis description cs:.5,-.05)},anchor=south},
		ylabel=Emotion Det. Accuracy (\%),
		label style={font=\tiny},
		ticklabel style = {font=\tiny},
		y tick label style={
			/pgf/number format/.cd,
			fixed,
			fixed zerofill,
			precision=0,
			/tikz/.cd
		},
		]
		\addplot[color=blue, thick, dashed] table{simple_facial_conv5_rank.txt}; \addlegendentry{noisy reduced simple}
		\addplot[color=awesome, thick] table{siamese_facial_conv5_rank.txt}; \addlegendentry{advanced}
		\end{axis}
		\end{tikzpicture}
		\caption{Comparison of presence or absence of Siamese fine-tunning.}
		\label{fig:emotion_presence}	
	\end{subfigure}
\caption{Transfer learning and Accuracy-Privacy trade-off on emotion detection, using VGG-S architecture and Conv5 as the intermediate layer. }
\end{figure}

\subsubsection{Transfer learning}
We test different embeddings with the transfer learning and the result are shown in Figure~\ref{fig:FRED}. The accuracy of the face recognition model is decreased for all embeddings. Similar to the gender classification application, the Siamese embedding works better than simple embedding and dimensionality reduction helps with privacy protection. The effect of different embeddings on emotion detection are reported in Table~\ref{tab:ed}. It is evident that the Siamese embedding does not decrease emotion detection accuracy significantly, while dimensionality reduction has  major impact on this task.

\begin{table}[t]
	\caption{Comparison of Different Emotion Detection Models. Intermediate Layer is Conv5.}
	\begin{center}
		\begin{tabular}{ |c|c| }
			\hline
			\multicolumn{2}{|c|}{Accuracy on SFEW-2} \\
			\hline
			simple \cite{levi2015emotion}  &  40\% \\
			Siamese & 38\% \\
			reduced simple  & 31\% \\
			reduced Siamese  & 32\% \\
			\hline
		\end{tabular}
	\end{center}
	\label{tab:ed}
\end{table}


\subsubsection{Privacy metric}
The results of the feature extractor validation are shown in Figure~\ref{fig:emotion_presence}, where the advanced embedding curve is above the noisy reduced simple curve. By having a fixed accuracy level, we can have higher privacy for advanced embedding. 

Results of the both applications show that our framework is application, and model, independent. The Siamese structure improves privacy, while reducing the dimensionality does not hurt the $CT_1$ accuracy and lowers the communication cost. We can use the validation method to quantify the privacy level, without access to the cloud-based face recognition model.

%

\subsection{Mobile Evaluation}

In the previous sections we presented different solutions for learning inferences. Cloud based solutions are robust, but do not respect the users' privacy. On the other hand, on-premise solutions have increased level of privacy but are not power efficient, decreasing the battery life of each mobile device. In this section we evaluate a new, hybrid approach, that is based on the methods explained in the previous sections. By reducing the complexity of the deep neural network, we managed to also reduce the loading time, inference time and memory usage, while at the same time hide the user's sensitive information.

\begin{figure}[t]
	\begin{center}
		\includegraphics[width=\linewidth]{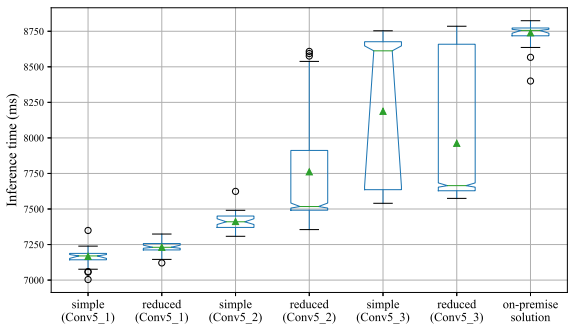} 	
	\end{center}
	\caption{Inference time of different deep embeddings on mobile (60 inferences per configuration).}
	\label{fig:mobile_time}
\end{figure}
\begin{figure}[t]
	\begin{center}
		\includegraphics[width=\linewidth]{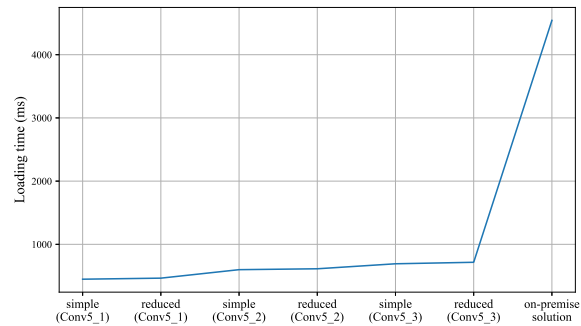} 	
	\end{center}
	\caption{Loading time comparison of different deep embeddings on mobile.}
	\label{fig:mobile_loading_time}
\end{figure}
\begin{figure}[t]
	\begin{center}
		\includegraphics[width=\linewidth]{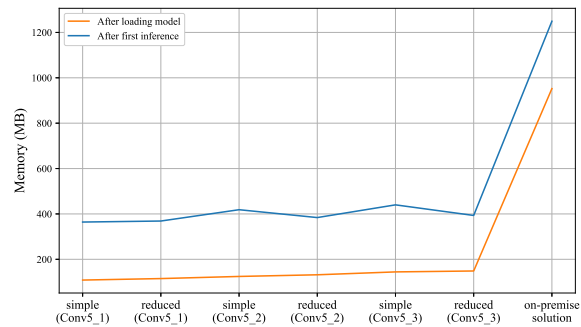} 	
	\end{center}
	\caption{Memory comparison of different deep embeddings on mobile.}
	\label{fig:mobile_memory}
\end{figure}

We evaluated the proposed implementation on a modern handset device, shown in Table~\ref{tab:device_spec}. In order to have a better comparison, we focus on the gender classification VGG-16 architecture and. We evaluated each solution separately (simple, reduced) for each of the three intermediate layers (Conv5\_1, Conv5\_2, Conv5\_3), and compared them with the on-premise solution (full model). We used Caffe Mobile~v1.0~\cite{jia2014caffe} for Android to load each model and measured the inference time (Figure~\ref{fig:mobile_time}), model loading time (Figure~\ref{fig:mobile_loading_time}) and model memory usage (Figure~\ref{fig:mobile_memory}) of each of the seven configurations. We configured the model to only use one core of the CPU, as the aim of this experiment was a comparison between the different techniques on the specific device.

\begin{table}
\centering
\caption{Device Specification}
\label{tab:device_spec}
\begin{tabular}{| l | l |}
	\hline
	\multicolumn{2}{|c|}{Google (Huawei) Nexus 6P} \\
    \hline
    Memory        & 3 GB LPDDR4 RAM \\
	Storage       & 32 GB \\
    CPU           & Octa-core Snapdragon 810 v2.1  \\
    GPU           & Adreno 430 \\
    OS            & Android 7.1.2 \\
    \hline
\end{tabular}
\end{table}

Most of the variations of trained model architectures under the proposed embedding approach report the same loading time and memory usage performance. There is a large increase in both memory use (217.66\%) and loading time (534.49\%) when loading the on-promise solution, proving the efficiency of our solution. Inference time also increases per configuration due to the increased size of the model.

We conclude that our approach is feasible to be implemented in a modern smartphone. By choosing a privacy-complexity trade-off and using different intermediate layers we were able to significantly reduce the cost when running the model on the mobile device, while at the same time preserving important user information from being uploaded to the cloud.

\section{Related Work}\label{sec:related}

In this section, we describe the prior works on privacy-preserving learning systems and their intrinsic differences. We also review the works used deep learning on mobile phones.

\subsection{Learning with privacy}

Prior works have approached the problem of privacy in machine learning from different point of views. Some approaches attempt to remove the irrelevant information by increasing the amount of uncertainty, while others try to hide information using cryptographic operations. Early works in this space mainly focus on publishing datasets for learning tasks~\cite{agrawal2000, agrawal2001, sweeney2002, iyengar2002}. They usually concern about publishing a dataset consists of high level features for data mining tasks (e.g., medical database consisting of patients details), while preserving the individuals' privacy. Solutions such as randomized noise addition~\cite{agrawal2000, agrawal2001} and k-anonymity by generalization and suppression \cite{lefevre2005, machanavajjhala2007, li2007} are proposed and surveyed in \cite{aggarwal2008}. These methods have some major problems. They are just appropriate for low-dimensional data due to the curse of dimensionality~\cite{aggarwal2005}, hence they are not fit most of the multimedia data. Also a variety of attacks make many these methods unreliable~\cite{aggarwal2008}. We can categorize these models as the \emph{dataset publishing} models. In dataset publishing, training applicability of a generalized data is important, while in this paper we deal with the cases where model training has been done already by a cloud service (e.g., Facebook or Google using their image data).

Differential privacy~\cite{dwork2008} is another method provides an exact way to publish statistics of a database while keeping all individual records of the database private. A learning model trained on some dataset can be considered as a high level statistic of that dataset. So considering the training data privacy while publishing a learning model is another important problem, we call it \emph{model publishing.} Recently, \cite{shokri2015} proposed concern of privacy for deep learning and \cite{abadi2016} provided differential private deep learning model. In \emph{model publishing}, mainly the privacy of users participating in training data is of concern, while in our scenario, user's data may not exist in training data and we focus on inference phase of a learning model. 

Hence, neither publishing a learning dataset or a learning model are directly relevant to our problem. We can name our problem as the \emph{secure inference} where the user can not access the learning model during inference time and should use it in a secure manner. A popular approach to solve this problem is reliance on cryptography methods. In~\cite{avidan2006}, the authors provide a secure protocol for machine learning. In~\cite{gilad2016}, the neural network is held in cloud. They encrypt the input of neural network in a way that inference becomes applicable on encrypted message. This approach has important, yet highly complex operations, making it infeasible. Mainly, the throughput is the same for inference on a single image or a batch. In addition neural network should be changed in a complex way to enable homomorphic encryption taking 250 seconds on a PC, which makes it impractical in terms of usability on a mobile phones or simple PCs. Recently \cite{rouhani2017deepsecure, mohassel2017secureml} tried to improve this work by implying a mored advance encryption setting, while they are still using simple deep models in experiments. 

Instead of encryption-based methods, we recommend a new approach to this problem, which is a kind of feature extraction, applied in a hybrid framework. We address this issue in an adversarial setting. We optimize a cost function which consist of data privacy and model accuracy terms. We then use the Siamese architecture to solve this optimization and get the private feature which is non-informative about sensitive information and can be shared with the cloud service.

 
\subsection{Privacy in image analytics}

Privacy preservation has also been addressed in machine vision community. A good survey of all methods attempted to provide visual privacy, can be found in \cite{padilla2015}, which classifies different methods to five categories: intervention, blind vision, secure processing, redaction and data hiding. Our work is similar  in spirit to de-identification works, a subcategory of redaction methods. The goal of these methods is to purturbe the individuals' faces in images in such a way that they can not be recognized by a face recognition system. A fundamental work in this category is presented in~\cite{newton2005}, which targets privacy issue in video surveillance data. The aim of this work is to publish a transformed dataset, where individuals are not identifiable. They show that using simple image filtering can not guarantee privacy and suggest \textit{K-same} algorithm, based on k-anonymity, aiming to create average face images and replace them with the original ones. A shortcoming of this work is the lack of protection against future analyses on the dataset. Lots of works followed this idea and tried to improve it, mainly with the goal of publishing a dataset that is different from us. Their goal is not to protect privacy of a new face image, which is our concern. Follow-up works aim to transform a face image in a way that it is unrecognizable, while other analytics such as gender classification is possible. Most of the works in this area use visual filters or morphing to make the image unrecognizable~\cite{korshunov2013, rachaud2015}. One of the main issues with prior privacy preservation methods is the lack of a privacy guarantee against new models due to engineering features against specific learning tasks. In most cases the learning task is not explicitly defined. Moreover, many works ignore the accuracy constraints of the learning task in their privacy preservation method. In this paper we build on our previous work~\cite{DBLP:journals/corr/OssiaSTRLH17}, introduce and develop a privacy measure, and evaluate the framework on smartphones.


\subsection{Deep learning on mobile phone}


Last two years have seen a dramatic increase in the implementation and inference ability of deep neural networks on smartphones. Using pre-trained deep learning models can increase accuracy of different sensors; e.g. in~\cite{lane2015can}, Lane \etal use a 3 layer network which does not overburden the hardware. Complex networks with more layers need more processing power. Architectures such as the 16-layer model (VGG-16) proposed in \cite{simonyan2014very} and the 8-layer model (\textit{VGG-S}) proposed in  \cite{chatfield14} which are more complex, are implemented on the mobile in~\cite{kim2015compression}, and the resource usage such as time, CPU and energy overhead, are reported. As most of the state of the art models are pretty large in scale, fully evaluating all the layers on mobile results in serious drawbacks in processing time and memory requirements. Some methods are proposed to approximate these complex functions with simpler ones to reduce the cost of inference. Kim \etal~\cite{kim2015compression} aim to compress deep models and in~\cite{bhattacharya2016} the authors use sparsification and kernel separation. However, the increase in efficiency of these methods comes with a decrease in accuracy of the model. In order to get more efficient results, we can also implement models on GPU. An implementation on GPU in~\cite{kim2015compression} has burdens on the battery, hence it is not a feasible solution for some practical applications that either users frequently use it or continuously require it for long periods \cite{lane2016deepx}. On the other hand, recent devices have DSP modules though their capacity for programming and storage can be limited. To tackle these problems, Lane~\etal~\cite{lane2016deepx} have implemented a software accelerator called DeepX for large-scale deep neural networks to reduce the resources while the mobile is doing inference by using different kinds of mobile processor simultaneously.


\section{Discussions and next steps}\label{sec:conclusions}

In this paper, we presented a new hybrid framework for efficient privacy preserving analytics which consists of a feature extractor and analyzer, where the former is placed on the client side and the later on the server side. We embed deep neural networks, specially, Convolutional neural networks in this framework to benefit from their accuracy and layered architecture. In order to protect the data privacy against unauthorized tasks, we used the Siamese architecture, creating a feature which is specific to the desired task. This is in contrast to today's ordinary deep networks in which the created features are generic and can be used for different tasks. Removing the undesired sensitive information from the extracted feature results in achieving privacy for the user. Evaluating our framework by splitting the layers between the mobile and the cloud and by targeted noise addition, we achieved high accuracy on the primary tasks, while heavily decreasing any inference potential for other tasks. Also by implementing the framework on mobile phone, we show that we can highly decrease the computational complexity on the user side, as well as the communication cost. 

Our framework is currently designed for learning inferences in the test phase. In ongoing work we are extending our method by designing a framework for Learning as a Service, where the users could share their data, in a privacy-preserving manner, to train a new learning model. Another potential extension to our framework will be providing support for other kinds of neural networks such as recurrent neural network and also other applications for speech or video processing. 

\section*{Acknowledgment} 
We would like to thank Sina Sajadmanesh for his valuable comments and feedbacks.


\bibliographystyle{IEEEtran}
%


\end{document}